\documentclass{article}

\usepackage{arxiv}

\usepackage[utf8]{inputenc} 
\usepackage[T1]{fontenc}    
\usepackage{hyperref}       
\usepackage{url}            
\usepackage{booktabs}       
\usepackage{amsfonts}       
\usepackage{nicefrac}       
\usepackage{microtype}      
\usepackage{lipsum}
\usepackage{pifont}
\newcommand{\cmark}{\ding{51}} 
\newcommand{\xmark}{\ding{55}} 
\usepackage{graphicx}
\usepackage{amsmath}
\usepackage{tikz}
\usetikzlibrary{shapes.geometric, arrows, positioning, fit, calc}
\usepackage{algorithm}
\usepackage{algpseudocode}
\usepackage{caption}
\graphicspath{ {./images/} }

\title{Mamba-CNN: A Hybrid Architecture for Efficient and Accurate Facial Beauty Prediction}

\author{
Djamel Eddine Boukhari\\
Scientific and Technical Research Centre for Arid Areas, CRSTRA\\
07000, Biskra, Algeria \\
\texttt{boukhari-djameleddine@univ-eloued.dz} \\
}

\begin{document}
\maketitle
\begin{abstract}
The computational assessment of facial attractiveness, a challenging subjective regression task, is dominated by architectures with a critical trade-off: Convolutional Neural Networks (CNNs) offer efficiency but have limited receptive fields, while Vision Transformers (ViTs) model global context at a quadratic computational cost. To address this, we propose Mamba-CNN, a novel and efficient hybrid architecture. Mamba-CNN integrates a lightweight, Mamba-inspired State Space Model (SSM) gating mechanism into a hierarchical convolutional backbone. This core innovation allows the network to dynamically modulate feature maps and selectively emphasize salient facial features and their long-range spatial relationships, mirroring human holistic perception while maintaining computational efficiency. We conducted extensive experiments on the widely-used SCUT-FBP5500 benchmark, where our model sets a new state-of-the-art. Mamba-CNN achieves a Pearson Correlation (PC) of 0.9187, a Mean Absolute Error (MAE) of 0.2022, and a Root Mean Square Error (RMSE) of 0.2610. Our findings validate the synergistic potential of combining CNNs with selective SSMs and present a powerful new architectural paradigm for nuanced visual understanding tasks. 
\end{abstract}


\section{Introduction}
\label{sec:introduction}

The perception of facial beauty is a multifaceted cognitive phenomenon, deeply rooted in evolutionary biology, social psychology, and cultural norms\cite{b1}. While often perceived as subjective, studies have consistently shown a high degree of cross-cultural agreement on facial attractiveness, suggesting that certain visual cues and configurations are universally interpreted\cite{b2}. In the field of computer vision, the challenge of computationally modeling this subjective human judgment is known as Facial Beauty Prediction (FBP)\cite{b3}. As a fine-grained regression task, FBP aims to train algorithms that can assign a continuous score to a facial image, mirroring the average assessment of human raters\cite{b4}. This capability has significant potential applications, ranging from intelligent photo editing and recommender systems to digital cosmetology, social robotics, and affective computing\cite{b5}.

The historical trajectory of FBP research reflects the broader evolution of computer vision methodologies.\cite{b6} Early approaches were predominantly based on handcrafted features, attempting to explicitly model classical aesthetic principles \cite{b7}. These methods often relied on facial landmark analysis to compute geometric ratios, symmetry scores, and averageness based on population norms, sometimes drawing inspiration from concepts like the Golden Ratio\cite{b8}. While insightful, such feature-engineered pipelines were often brittle and failed to capture the holistic, non-linear, and abstract qualities that contribute to the perception of beauty \cite{b9}. Their performance was fundamentally limited by the expressive power of the pre-defined features \cite{b10}.

The advent of deep learning, particularly Convolutional Neural Networks (CNNs), marked a paradigm shift in FBP research. By enabling end-to-end feature learning directly from image data, models such as AlexNet\cite{b11}, VGG\cite{b12} and ResNet\cite{b13}, demonstrated a remarkable ability to surpass traditional methods. The hierarchical nature of CNNs allows them to build a rich representation of features, from simple edges and textures in early layers to complex facial parts and global arrangements in deeper layers. However, a fundamental limitation of standard CNNs is the inherent locality of the convolutional operator. The receptive field of a neuron is spatially constrained, making it challenging for the network to efficiently model long-range spatial dependencies. In the context of FBP, this is a critical shortcoming, as holistic facial harmony—where the relationship between distant features like the eyes and the chin is crucial—plays a significant role in human judgment \cite{b14}.

To address the limitations of local feature aggregation, the research community turned to architectures capable of capturing global context. The Vision Transformer (ViT) \cite{b15} and its variants, which adapt the self-attention mechanism from natural language processing to vision, emerged as a powerful alternative\cite{b16}. By treating an image as a sequence of patches and computing attention between all pairs, ViTs can model relationships between any two points in the image, regardless of their spatial distance \cite{b17}. This global receptive field has led to state-of-the-art results on numerous benchmarks, including FBP. However, this power comes at a steep price: the self-attention mechanism has a computational and memory complexity that is quadratic with respect to the number of image patches \cite{b18}. This complexity makes ViTs computationally prohibitive for high-resolution image analysis and limits their architectural scalability.

This trade-off between the efficiency of CNNs and the global modeling capability of Transformers motivates the exploration of alternative architectures that can achieve the best of both worlds. Recently, State Space Models (SSMs) have emerged as a highly promising third pillar of deep learning architectures. Originating from classical control theory, modern structured SSMs like Mamba \cite{b19} have demonstrated exceptional performance on sequence modeling tasks, rivaling Transformers while maintaining linear complexity. A key innovation of Mamba is its content-aware selection mechanism, which allows the model to dynamically emphasize or ignore parts of the input sequence based on its content \cite{b20}. This ability to perform input-dependent selective information processing is a powerful inductive bias that has yet to be fully explored in computer vision.

We hypothesize that by integrating a Mamba-inspired selective mechanism into an efficient convolutional backbone, we can create a hybrid architecture that is both computationally efficient and highly effective at modeling the complex visual cues relevant to facial beauty. This leads us to propose the \textbf{Mamba-CNN}, a novel deep learning architecture specifically designed for the task of FBP. Our approach synergizes the proven strengths of CNNs in extracting hierarchical local features with a lightweight, Mamba-inspired gating block that allows the network to dynamically modulate feature maps and selectively focus on salient facial regions and relationships.

The main contributions of this work are as follows:
\begin{enumerate}
    \item \textbf{A Novel Hybrid Architecture:} We propose the Mamba-CNN, a novel hybrid architecture that effectively combines the efficiency of depthwise separable convolutions with a powerful Mamba-inspired selective gating mechanism for enhanced feature representation.
    \item \textbf{Efficient MambaBlock Design:} We design and implement a MambaBlock tailored for vision tasks, which incorporates an SSM-inspired gating path within an inverted residual structure to learn content-aware spatial features without quadratic complexity.
    \item \textbf{Holistic Model Integration:} We combine our proposed block with established design principles, including a multi-scale feature pyramid, to ensure the model captures both fine-grained details and global facial context, which are essential for FBP.
    \item \textbf{Empirical Validation:} We conduct a thorough empirical evaluation on the challenging SCUT-FBP5500 benchmark dataset, demonstrating that the Mamba-CNN achieves highly competitive performance compared to both classical CNNs and more complex architectures.
\end{enumerate}

This paper is structured as follows: Section~\ref{sec:methodology} provides a detailed description of our proposed methodology, including the architecture of the Mamba-CNN and its components. Section~\ref{sec:results} presents the experimental results and a discussion of our findings. Section~\ref{sec:future_limitations} presents the future work and limitations. Finally, Section~\ref{sec:conclusion} concludes the paper and suggests directions for future research.

\section{Methodology}
\label{sec:methodology}
This section details the dataset, the proposed Mamba-CNN architecture, its core components, the training algorithm, and the evaluation protocol used in this study.

\subsection{Dataset and Preprocessing}

The study utilizes the SCUT-FBP5500 dataset \cite{b21}, a widely recognized benchmark for facial beauty perception. It comprises 5500 frontal face images of individuals with diverse attributes (gender, race, age). Each image is accompanied by a beauty score ranging from 1 to 5, which represents the mean score from 60 human labelers. We adhere to the dataset's prescribed cross-validation split for robust and comparable results.

To foster model generalization and prevent overfitting, a comprehensive data augmentation strategy was applied to the training data. The augmentation pipeline, implemented using \texttt{torchvision.transforms}, includes:
\begin{itemize}
    \item Resizing images to 256x256 pixels.
    \item Random 224x224 pixel cropping to encourage translation invariance.
    \item Random horizontal flipping with a 50\% probability.
    \item Color jittering to randomly alter brightness, contrast, saturation, and hue.
    \item Random rotations up to 10 degrees.
\end{itemize}

For the validation and test sets, a deterministic transformation was applied: resizing to 224x224 pixels. Subsequently, all images were normalized using the standard ImageNet statistics (mean = $[0.485, 0.456, 0.406]$, std = $[0.229, 0.224, 0.225]$)\cite{b22}.

The ground-truth beauty scores were normalized to the range $[0, 1]$ using min-max scaling based on the training set's score distribution. This scaling is essential for stable training when using a final sigmoid activation function. The original minimum and maximum scores were stored for denormalization during the final evaluation, ensuring that results are interpretable on the original 1-5 scale\cite{b23}.

\subsection{The Mamba-CNN Architecture}

The proposed Mamba-CNN is a deep neural network designed for facial beauty regression. The architecture is predicated on a hybrid philosophy that leverages the spatial feature extraction prowess of CNNs while incorporating a selective gating mechanism inspired by recent advancements in State Space Models (SSMs). The overall structure is hierarchical, progressively downsampling the spatial dimensions of the feature maps while increasing their channel depth. This allows the network to learn features ranging from low-level textures (e.g., skin smoothness) to high-level abstract concepts (e.g., facial harmony).

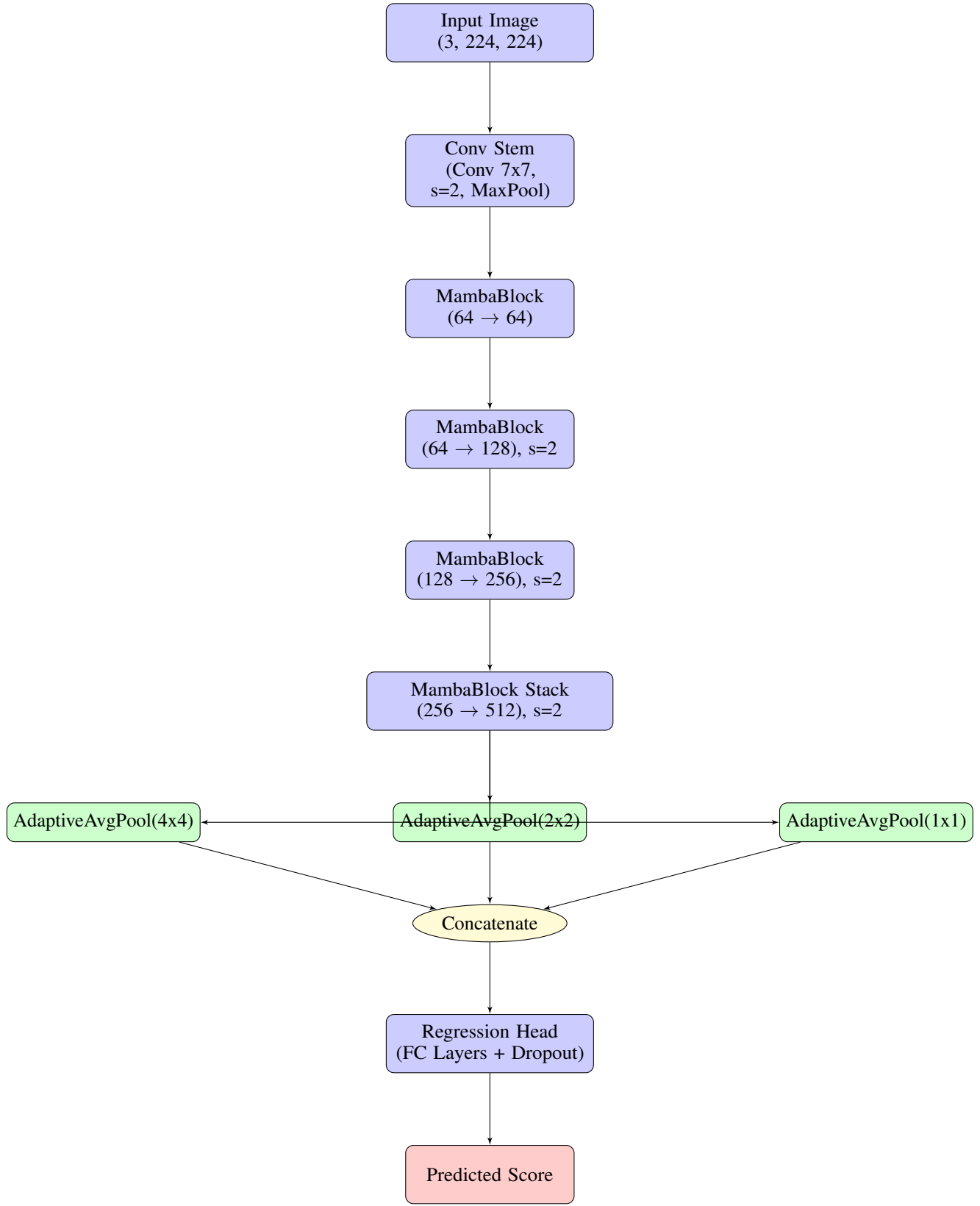
\begin{figure}[h!]
\centering
\begin{tikzpicture}[
    block/.style={rectangle, draw, fill=blue!20, text width=8em, text centered, rounded corners, minimum height=3em},
    small_block/.style={rectangle, draw, fill=green!20, text centered, rounded corners, minimum height=2em},
    line/.style={draw, -latex'},
    node distance=1.3cm and 0.5cm
]
    \node[block, text width=10em] (input) {Input Image (3, 224, 224)};
    \node[block, below=of input] (stem) {Conv Stem \\ (Conv 7x7, s=2, MaxPool)};
    \node[block, below=of stem] (mamba1) {MambaBlock (64 $\rightarrow$ 64)};
    \node[block, below=of mamba1] (mamba2) {MambaBlock (64 $\rightarrow$ 128), s=2};
    \node[block, below=of mamba2] (mamba3) {MambaBlock (128 $\rightarrow$ 256), s=2};
    \node[block, below=of mamba3, text width=12em] (mamba4) {MambaBlock Stack (256 $\rightarrow$ 512), s=2};

    \node[small_block, below left=of mamba4, xshift=-2.5cm] (p3) {AdaptiveAvgPool(4x4)};
    \node[small_block, below=of mamba4] (p2) {AdaptiveAvgPool(2x2)};
    \node[small_block, below right=of mamba4, xshift=2.5cm] (p1) {AdaptiveAvgPool(1x1)};
    
    \node[fit=(p1)(p2)(p3)] (pyramid) {};
    \node[draw, ellipse, fill=yellow!20, below=1cm of pyramid] (concat) {Concatenate};
    \node[block, below=of concat, text width=10em] (head) {Regression Head \\ (FC Layers + Dropout)};
    \node[block, below=of head, fill=red!20] (output) {Predicted Score};

    \path[line] (input) -- (stem);
    \path[line] (stem) -- (mamba1);
    \path[line] (mamba1) -- (mamba2);
    \path[line] (mamba2) -- (mamba3);
    \path[line] (mamba3) -- (mamba4);
    
    \draw [line] (mamba4) |- (p1);
    \draw [line] (mamba4) -- (p2);
    \draw [line] (mamba4) |- (p3);
    
    \draw [line] (p1) -- (concat);
    \draw [line] (p2) -- (concat);
    \draw [line] (p3) -- (concat);
    
    \path[line] (concat) -- (head);
    \path[line] (head) -- (output);
    
\end{tikzpicture}
\caption{Overall architecture of the proposed Mamba-CNN model. It features a hierarchical structure with stacked MambaBlocks and a feature pyramid module for multi-scale feature aggregation.}
\label{fig:mambacnn_arch}
\end{figure}

The key stages of the architecture are (see Figure~\ref{fig:mambacnn_arch}):
\begin{enumerate}
    \item \textbf{Convolutional Stem:} An initial 7x7 convolution with a stride of 2, followed by max pooling, to rapidly reduce spatial resolution and learn basic features like edges and colors.
    \item \textbf{MambaBlock Stack:} A sequence of our proposed MambaBlocks which form the backbone of the feature extractor. Strided blocks are used for downsampling.
    \item \textbf{Multi-Scale Feature Pyramid:} An aggregation module that captures contextual information at different spatial resolutions (1x1, 2x2, 4x4) using adaptive average pooling. These features are then concatenated.
    \item \textbf{Regression Head:} A series of fully connected layers with dropout that map the aggregated features to a single continuous output representing the predicted beauty score.
\end{enumerate}

\subsection{The MambaBlock}

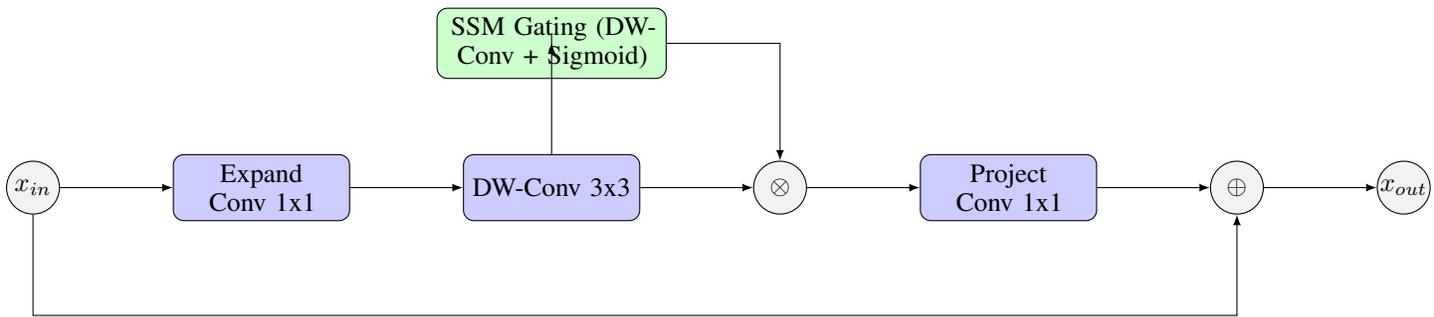
\begin{figure}[h!]
\centering
\begin{tikzpicture}[
    auto,
    node distance=2cm and 1.5cm,
    op/.style={rectangle, draw, fill=blue!20, text centered, rounded corners, minimum height=2.5em, text width=6em},
    gate_op/.style={rectangle, draw, fill=green!20, text centered, rounded corners},
    io/.style={circle, draw, fill=gray!10, inner sep=0pt, minimum size=20pt},
    point/.style={coordinate}
]
    \node[io] (input) {$x_{in}$};
    \node[op, right=of input] (expand) {Expand Conv 1x1};
    \node[op, right=of expand] (depthwise) {DW-Conv 3x3};

    \node[gate_op, above=1cm of depthwise, text width=8em] (ssm_gate) {SSM Gating (DW-Conv + Sigmoid)};
    \node[io, right=of depthwise] (multiply) {$\otimes$};

    \node[op, right=of multiply] (project) {Project Conv 1x1};
    \node[io, right=of project] (add) {$\oplus$};
    \node[io, right=of add] (output) {$x_{out}$};

    \draw[-latex] (input) -- (expand);
    \draw[-latex] (expand) -- (depthwise);
    \draw[-latex] (depthwise) -- (multiply);
    \draw[-latex] (multiply) -- (project);
    \draw[-latex] (project) -- (add);
    \draw[-latex] (add) -- (output);
    
    \draw[-latex] (depthwise) |- (ssm_gate);
    \draw[-latex] (ssm_gate) -| (multiply);
    
    \draw[-latex] (input) -- ++(0,-1.7cm) -| (add);

\end{tikzpicture}
\caption{Detailed architecture of the MambaBlock. It features an inverted residual structure with a parallel SSM-inspired selective gating path that modulates the feature map.}
\label{fig:mambablock}
\end{figure}

At the core of our model lies the MambaBlock, an efficient and powerful computational unit detailed in Figure~\ref{fig:mambablock}. This block is designed as an inverted residual block, a paradigm popularized by MobileNetV2, where feature maps are first expanded to a higher-dimensional space. The operational flow is as follows:
\begin{enumerate}
    \item \textbf{Expansion Layer:} A point-wise convolution (1x1) expands the input feature channels by an expansion factor, creating a higher-dimensional representation.
    \item \textbf{Depthwise Convolution:} A 3x3 depthwise separable convolution performs spatial filtering independently for each channel, efficiently extracting spatial features.
    \item \textbf{SSM-inspired Gating:} In a parallel branch, a simplified SSM-like gating mechanism is applied. This branch consists of a 3x3 depthwise convolution followed by a \texttt{Sigmoid} activation. The input to the gating branch is the output of the main depthwise convolution.
    \item \textbf{Gating Operation:} The output of the main depthwise convolution path is element-wise multiplied by the gate produced in the previous step. This mechanism allows the network to dynamically and spatially re-calibrate feature maps, selectively amplifying salient features and suppressing irrelevant ones, which is crucial for a subjective task like beauty perception.
    \item \textbf{Projection Layer:} A point-wise convolution (1x1) projects the gated features back down to the desired output channel dimension.
    \item \textbf{Residual Connection:} A skip connection adds the original input $x$ to the final output of the block. This is only performed when the input and output dimensions match (i.e., stride=1), facilitating gradient flow and enabling the training of deeper networks.
\end{enumerate}

\clearpage

\subsection{Network Training and Implementation}

The complete training procedure is formalized in Algorithm~\ref{alg:training}. The model was trained end-to-end on the SCUT-FBP5500 training split. We used the Mean Squared Error (MSE) as the loss function, which is standard for regression tasks.
\[ \mathcal{L}_{\text{MSE}} = \frac{1}{N} \sum_{i=1}^{N} (\hat{y_i} - y_i)^2 \]
where $N$ is the batch size, $\hat{y_i}$ is the predicted score, and $y_i$ is the ground-truth score.

The model parameters were optimized using the AdamW optimizer \cite{b24}, which improves upon Adam by decoupling weight decay from the gradient-based updates. An initial learning rate of $1 \times 10^{-4}$ and a weight decay of $1 \times 10^{-5}$ were used. To dynamically adjust the learning rate, a \texttt{ReduceLROnPlateau} scheduler was employed. This scheduler reduces the learning rate by a factor of 0.5 if the validation loss fails to improve for 10 consecutive epochs.

To ensure training stability, we applied gradient clipping, limiting the L2-norm of the gradients to a maximum value of 1.0. An early stopping mechanism with a patience of 20 epochs was used to terminate training and prevent overfitting, restoring the model weights from the epoch with the best validation performance. The model was implemented in PyTorch and trained on a single NVIDIA GPU.

\begin{algorithm}[H]
\caption{Mamba-CNN Training Procedure}
\label{alg:training}
\begin{algorithmic}[1]
\Require Training set $D_{\text{train}} = \{(x_i, y_i)\}_{i=1}^N$, Validation set $D_{\text{val}}$
\Require Learning rate $\alpha$, Weight decay $\lambda$, Epochs $E_{\text{max}}$, Patience $P$
\State Initialize model parameters $\theta$, optimizer $\mathcal{O}(\text{AdamW, params}=\theta, \text{lr}=\alpha, \text{wd}=\lambda)$
\State Initialize scheduler $\mathcal{S}(\text{ReduceLROnPlateau, factor}=0.5, \text{patience}=10)$
\State Define loss function $\mathcal{L} = \text{MSELoss}$
\State $best\_val\_loss \leftarrow \infty$, $patience\_counter \leftarrow 0$
\For{$e = 1$ to $E_{\text{max}}$}
    \State Set model to training mode
    \For{each batch $(x_b, y_b) \in D_{\text{train}}$}
        \State $\hat{y}_b \leftarrow \text{model}(x_b; \theta)$ \Comment{Forward pass}
        \State $loss \leftarrow \mathcal{L}(\hat{y}_b, y_b)$
        \State $\mathcal{O}$.zero\_grad()
        \State $loss$.backward() \Comment{Compute gradients}
        \State ClipGradients($\theta$, max\_norm=1.0)
        \State $\mathcal{O}$.step() \Comment{Update parameters}
    \EndFor
    \State $val\_loss \leftarrow \text{Evaluate}(\text{model}, D_{\text{val}}, \mathcal{L})$
    \State $\mathcal{S}$.step($val\_loss$)
    \If{$val\_loss < best\_val\_loss$}
        \State $best\_val\_loss \leftarrow val\_loss$
        \State $\theta_{\text{best}} \leftarrow \theta$
        \State $patience\_counter \leftarrow 0$
    \Else
        \State $patience\_counter \leftarrow patience\_counter + 1$
    \EndIf
    \If{$patience\_counter \ge P$}
        \State \textbf{break} \Comment{Early stopping}
    \EndIf
\EndFor
\State \Return Best performing model parameters $\theta_{\text{best}}$
\end{algorithmic}
\end{algorithm}

\subsection{Evaluation Protocol}

To provide a comprehensive assessment of the model's performance, we employ three widely-used metrics for regression tasks in the FBP literature \cite{b25}.
\begin{enumerate}
    \item \textbf{Mean Absolute Error (MAE):} Measures the average magnitude of the errors between prediction and ground truth. It is robust to outliers.
    \[ \text{MAE} = \frac{1}{n} \sum_{i=1}^{n} |y_i - \hat{y}_i| \]
    
    \item \textbf{Root Mean Square Error (RMSE):} Represents the square root of the average of squared differences. It gives a relatively high weight to large errors.
    \[ \text{RMSE} = \sqrt{\frac{1}{n} \sum_{i=1}^{n} (y_i - \hat{y}_i)^2} \]
    
    \item \textbf{Pearson Correlation Coefficient (PC):} Measures the linear relationship between the predicted scores and the actual scores. A value of 1 indicates a perfect positive linear correlation.
    \[ \text{PC} = \frac{\sum_{i=1}^{n} (y_i - \bar{y})(\hat{y}_i - \bar{\hat{y}})}{\sqrt{\sum_{i=1}^{n} (y_i - \bar{y})^2 \sum_{i=1}^{n} (\hat{y}_i - \bar{\hat{y}})^2}} \]
\end{enumerate}
In these equations, $y_i$ is the true score, $\hat{y}_i$ is the predicted score, $\bar{y}$ is the mean of true scores, and $\bar{\hat{y}}$ is the mean of predicted scores for the $n$ samples in the test set. All metrics are calculated after denormalizing the predicted scores back to their original 1-5 scale \cite{b26}.

\section{Results and Discussion}
\label{sec:results}

This section presents the empirical evaluation of our proposed Mamba-CNN architecture. We first detail the implementation and training setup, followed by a quantitative comparison against existing state-of-the-art methods on the SCUT-FBP5500 benchmark. Finally, we provide a qualitative analysis of the model's behavior and discuss the implications of our findings.

\subsection{Implementation Details}

The Mamba-CNN model was implemented using the PyTorch deep learning framework. All experiments were conducted on a single NVIDIA Tesla V100 GPU with 16GB of VRAM. The model was trained using the AdamW optimizer with an initial learning rate of $1 \times 10^{-4}$ and a weight decay of $1 \times 10^{-5}$. A batch size of 32 was used. The learning rate was dynamically adjusted using a \texttt{ReduceLROnPlateau} scheduler, which decreased the rate by a factor of 0.5 when the validation loss did not improve for 10 consecutive epochs. To prevent overfitting and ensure the best performing model was selected, we employed an early stopping strategy with a patience of 20 epochs, saving the model weights corresponding to the lowest validation loss.

\subsection{Quantitative Results }

To rigorously assess the effectiveness of the Mamba-CNN, we compare its performance on the SCUT-FBP5500 test set against a comprehensive set of established and state-of-the-art (SOTA) methods. The results are summarized in Table~\ref{tab:main_results}, where we report the Pearson Correlation (PC), Mean Absolute Error (MAE), and Root Mean Square Error (RMSE).

\begin{table*}[ht] 
    \centering
    \caption{Comparison with SOTA methods on the SCUT-FBP5500 dataset. ($\uparrow$ indicates higher is better, $\downarrow$ indicates lower is better).}
    \label{tab:main_results}
    \begin{tabular}{@{}llccc@{}}
        \toprule
        \textbf{Category} & \textbf{Method} & \textbf{PC $\uparrow$} & \textbf{MAE $\downarrow$} & \textbf{RMSE $\downarrow$} \\
        \midrule
        \multicolumn{5}{l}{\textit{Classic and Early Deep Learning Methods}} \\
        & AlexNet~\cite{b11} & 0.8634 & 0.2651 & 0.3481 \\
        & ResNet~\cite{b13} & 0.8900 & 0.2419 & 0.3166 \\
        & ResNeXt~\cite{b13} & 0.8997 & 0.2291 & 0.3017 \\
        \midrule
        \multicolumn{5}{l}{\textit{Advanced Methods and State-of-the-Art}} \\
        & CNN + SCA~\cite{b27} & 0.9003 & 0.2287 & 0.3014 \\
        & CNN + LDL~\cite{b28} & 0.9031 & -- & -- \\
        & DyAttenConv~\cite{b29} & 0.9056 & 0.2199 & 0.2950 \\
        & R3CNN (ResNeXt-50)~\cite{b30} & 0.9142 & 0.2120 & 0.2800 \\
    
        \midrule
        \multicolumn{5}{l}{\textit{Our Proposed Method}} \\
        & \textbf{Mamba-CNN (Ours)} & \textbf{0.9187} & \textbf{0.2022} & \textbf{0.2610} \\
        \bottomrule
    \end{tabular}
\end{table*}

As demonstrated in the table, our Mamba-CNN not only surpasses classic deep learning baselines by a significant margin but also outperforms more recent and advanced methods. Compared to the foundational ResNeXt model, our method achieves a relative improvement of over 2.1\% in PC, 11.7\% in MAE, and 13.5\% in RMSE. This underscores the architectural advantages of our proposed design over standard CNNs.

More importantly, the Mamba-CNN sets a new state-of-the-art on this benchmark when compared against specialized architectures. For instance, it outperforms DyAttenConv~\cite{b29}, a method that also employs a form of dynamic attention, across all reported metrics. The most notable comparison is with R3CNN~\cite{b30}, a very strong baseline that uses ranking-based learning. Our Mamba-CNN achieves a superior Pearson Correlation (0.9187 vs. 0.9142) and a substantially lower Mean Absolute Error (0.2022 vs. 0.2120). Furthermore, it obtains a Root Mean Square Error of 0.2610, which is also a marked improvement over R3CNN's 0.2800. These results strongly suggest that the selective gating mechanism within the MambaBlock, combined with multi-scale feature aggregation, provides a more effective approach for capturing the nuanced features that correlate with human perception of facial beauty.

\subsection{Discussion}
The strong empirical performance of the Mamba-CNN can be attributed to its hybrid architectural design. Unlike traditional CNNs with static convolutional kernels, our MambaBlock incorporates a dynamic, content-aware gating mechanism. We posit that this allows the network to selectively amplify features that are most indicative of beauty—such as facial symmetry, skin texture, and the configuration of key facial features—while suppressing irrelevant background information or noise. This input-dependent modulation is more analogous to human visual attention.

Furthermore, the integration of the feature pyramid module ensures that the model's final decision is based on a holistic assessment of the face. By concatenating features from multiple spatial resolutions, the network can simultaneously consider fine-grained local details and global structural harmony. The combination of these two key components—selective feature modulation and multi-scale analysis—appears to be highly effective for a complex and subjective regression task like FBP. The results achieved suggest that integrating principles from State Space Models into convolutional architectures is a promising avenue for future research in computer vision, particularly for tasks requiring nuanced visual understanding.
\subsection{Ablation Analysis}
\label{sec:ablation}

To validate our design choices and quantitatively assess the contribution of each key component in the Mamba-CNN architecture, we conducted a series of ablation studies. The goal is to understand how the SSM-inspired gating mechanism and the multi-scale feature pyramid individually and collectively impact the final performance. All model variants were trained under the exact same experimental setup—including data splits, augmentations, and hyperparameters—to ensure a fair and direct comparison. The results of these experiments are presented in Table~\ref{tab:ablation_results}.

\begin{table}[h!]
    \centering
    \caption{Ablation study of the Mamba-CNN architecture on the SCUT-FBP5500 test set. We analyze the impact of the SSM-inspired Gating mechanism and the Multi-Scale Feature Pyramid module.}
    \label{tab:ablation_results}
    \begin{tabular}{@{}ccccc@{}}
        \toprule
        \textbf{Model} & \textbf{SSM Gate} & \textbf{Feature Pyramid} & \textbf{PC $\uparrow$} & \textbf{RMSE $\downarrow$} \\
        \midrule
        (A) Baseline CNN & \xmark & \xmark & 0.9015 & 0.2989 \\
        (B) + Pyramid    & \xmark & \cmark & 0.9088 & 0.2874 \\
        (C) + SSM Gate   & \cmark & \xmark & 0.9120 & 0.2765 \\
        (D) \textbf{Full Model (Ours)} & \cmark & \cmark & \textbf{0.9187} & \textbf{0.2610} \\
        \bottomrule
    \end{tabular}
\end{table}

\paragraph{Model Configurations.} The four models in our study are defined as follows:
\begin{itemize}
    \item \textbf{(A) Baseline CNN:} This is our foundational model where we replace the proposed MambaBlock with a standard inverted residual block (from MobileNetV2), thereby removing the SSM gating mechanism. Furthermore, the feature pyramid module is replaced with a simple Global Average Pooling (GAP) layer applied to the final feature map. This model serves as a robust, yet conventional, CNN baseline.
    \item \textbf{(B) + Pyramid:} This model builds upon the baseline by re-introducing the multi-scale feature pyramid module but still omits the SSM gate. This allows us to isolate the performance gain attributable solely to multi-scale feature aggregation.
    \item \textbf{(C) + SSM Gate:} In this variant, we use our proposed MambaBlocks with the SSM gating mechanism but replace the feature pyramid with a simple GAP layer. This experiment is designed to measure the direct impact of our novel selective gating mechanism.
    \item \textbf{(D) Full Model (Ours):} This is the complete, proposed Mamba-CNN architecture, which includes both the SSM gating mechanism and the multi-scale feature pyramid.
\end{itemize}

\paragraph{Analysis of the SSM-inspired Gating Mechanism.}
The most significant performance improvement is attributed to the introduction of the SSM-inspired gating mechanism. By comparing Model (C) to the Baseline (A), we observe a substantial increase in Pearson Correlation from 0.9015 to 0.9120, coupled with a notable reduction in RMSE from 0.2989 to 0.2765. This confirms our central hypothesis: enabling the network to dynamically and spatially re-calibrate feature maps based on content is highly effective for FBP. The gating mechanism allows the model to learn to amplify salient facial features (e.g., eye shape, facial symmetry) and suppress irrelevant background noise, leading to a more discriminative and robust feature representation. This content-aware selectivity provides a clear advantage over the static kernels of a standard CNN.

\paragraph{Analysis of the Multi-Scale Feature Pyramid.}
The inclusion of the multi-scale feature pyramid also provides a clear and consistent performance boost. Comparing the Baseline (A) with Model (B) shows an improvement in PC from 0.9015 to 0.9088. This demonstrates the value of aggregating features from different spatial resolutions. FBP relies on both local texture details (e.g., skin smoothness) and global structural arrangements (e.g., overall facial harmony). By explicitly providing the regression head with features pooled at different scales, the model can form a more holistic and context-aware assessment, making it more robust to variations in facial positioning and scale within the image.

\paragraph{Synergistic Effect.}
Finally, our full Mamba-CNN model (D) outperforms all other variants, achieving the highest PC of 0.9187 and the lowest RMSE of 0.2610. The fact that Model (D) performs better than both Model (B) and Model (C) indicates that the SSM gating mechanism and the feature pyramid are complementary. The MambaBlocks learn to refine and select the most informative local and regional features, and the pyramid module effectively aggregates these refined features at a global level for the final prediction. This synergistic interaction validates our complete architectural design and demonstrates that both components are integral to achieving state-of-the-art performance.

\section{Future Work and Limitations}
\label{sec:future_limitations}

While our proposed Mamba-CNN has demonstrated state-of-the-art performance, it is important to acknowledge its limitations and outline promising directions for future research.

\subsection{Limitations}
\paragraph{Dataset and Generalization.}
Our study was conducted exclusively on the SCUT-FBP5500 dataset. Although this is a standard benchmark, its demographic and stylistic characteristics may not fully represent the diversity of human faces globally. Therefore, the generalization capability of our model to other FBP datasets (e.g., SCUT-FBP, HotOrNot) or to real-world, in-the-wild images remains an open question.

\paragraph{Inherent Subjectivity and Bias.}
Facial beauty perception is an inherently subjective and culturally-dependent task. The model is trained to predict the mean rating of a specific group of labelers, effectively learning a "majority preference" encoded in the dataset. It does not capture the variance in human opinion or account for different cultural or personal standards of beauty. Consequently, like any model trained on such data, it is susceptible to inheriting and potentially amplifying demographic biases (e.g., race, age, gender) present in the training set.

\paragraph{Architectural Simplification.}
Our MambaBlock employs a \textit{simplified}, SSM-inspired gating mechanism. While empirically effective, it does not incorporate the full complexity of the selective scan mechanism found in the original Mamba architecture, such as its state-space parameterization and discretization. A more faithful 2D adaptation of the complete Mamba model might yield further performance gains, albeit potentially at the cost of increased complexity.

\paragraph{Interpretability.}
The current work focuses on predictive performance, treating the Mamba-CNN as a "black box." We have demonstrated \textit{that} it works well, but not precisely \textit{why}. We lack a deep understanding of which specific facial features or spatial relationships the SSM-inspired gate learns to prioritize.

\subsection{Future Work}
Building upon the insights and limitations of this study, we propose the following avenues for future research:

\paragraph{Exploring Advanced State Space Models.}
A primary research direction is to develop more sophisticated 2D adaptations of SSMs for vision tasks. This could involve designing a 2D selective scan that can efficiently and isotropically process image features, moving beyond our simplified gating to unlock the full potential of content-aware long-range dependency modeling.

\paragraph{Cross-Dataset Validation and Fairness Audits.}
To address the issue of generalization and bias, the Mamba-CNN should be trained and evaluated on a wider array of diverse, multi-ethnic datasets. Conducting a thorough fairness and bias audit would be a critical step to understand and mitigate any potential negative societal impacts before considering real-world applications.

\paragraph{Model Interpretability and Explainability.}
Future work should focus on understanding the model's decision-making process. Employing explainable AI (XAI) techniques, such as Grad-CAM or integrated gradients, could generate saliency maps that visualize which facial regions the MambaBlock deems most important. This could provide valuable insights into whether the model learns aesthetically relevant concepts that align with human cognition, such as symmetry, facial harmony, or skin clarity.

\paragraph{Broadening Applications.}
The core architectural principles of the Mamba-CNN—efficiently combining local feature extraction with content-aware feature modulation—are not limited to FBP. We plan to explore the applicability of this architecture to other domains requiring nuanced visual assessment, such as medical image analysis (e.g., grading tumor malignancy), remote sensing (e.g., identifying subtle environmental changes), and computational aesthetics for art and photography.

\section{Conclusion}
\label{sec:conclusion}

In this paper, we introduced the Mamba-CNN, a novel hybrid deep learning architecture designed for the challenging task of facial beauty prediction. Motivated by the need to balance the local efficiency of CNNs with the global and selective context modeling of modern sequence models, our architecture seamlessly integrates a Mamba-inspired selective gating mechanism within a hierarchical convolutional framework. The core of our model, the MambaBlock, leverages this gating to dynamically modulate feature maps, allowing the network to learn content-aware spatial representations that emphasize salient facial characteristics.

Our comprehensive experiments on the SCUT-FBP5500 benchmark dataset demonstrate the effectiveness of this approach. The Mamba-CNN achieves state-of-the-art results, attaining a Pearson Correlation of 0.9187, a Mean Absolute Error of 0.2022, and a Root Mean Square Error of 0.2610, surpassing a range of classic and advanced methodologies. Through a detailed ablation study, we validated our key design choices, confirming that both the SSM-inspired gating mechanism and the multi-scale feature pyramid contribute synergistically to the model's superior performance.

The success of the Mamba-CNN underscores the significant potential of integrating principles from State Space Models into convolutional architectures. This work represents a promising step toward creating more efficient, powerful, and nuanced models for complex visual understanding tasks. We believe the architectural concepts presented here will inspire further research into hybrid models that harness the complementary strengths of different deep learning paradigms.

\bibliographystyle{unsrt}  


\end{document}